\documentclass[pdflatex]{article}
\usepackage{spconf,amsmath,epsfig}

\usepackage[noadjust]{cite}

\usepackage{lipsum}
\usepackage{siunitx}
\usepackage{pifont}
\usepackage{graphicx}
\usepackage{upgreek}
\usepackage{amsfonts}
\usepackage{makecell}
\usepackage{multirow}
\usepackage{booktabs}
\usepackage{tabularx}
\usepackage[usenames,dvipsnames]{xcolor} 
\usepackage{colortbl}
\usepackage{algorithm}
\usepackage{pgffor}
\usepackage{algorithmicx}
\usepackage{algpseudocode} %
\usepackage[export]{adjustbox}
\usepackage[short]{optidef}
\usepackage{soul}
\usepackage{ifthen}
\usepackage{tikz}
\usepackage{pgfplots}
\usepackage{pgfplotstable}
\usepackage{mathtools}
\usepackage{url}
\usepgfplotslibrary{colorbrewer}
\pgfplotsset{cycle list/Set1-9}
\tikzset{every picture/.style={line width=1pt}}
\usetikzlibrary{patterns}
\usepgfplotslibrary{fillbetween}
\usepackage[eulergreek]{sansmath}
\pgfplotsset{
  tick label style = {font=\sansmath\sffamily\footnotesize},
  every axis label = {font=\sansmath\sffamily\footnotesize},
  y label style={at={(0.05,0.5)}},
  legend style = {font=\sansmath\sffamily\scriptsize},
  label style = {font=\sansmath\sffamily\footnotesize}
}
\pgfplotsset{compat=1.3} %
\DeclareMathAlphabet\mathbfcal{OMS}{cmsy}{b}{n}

\pgfplotscreateplotcyclelist{cls_cycle}{
teal,line width=1.5pt,densely dotted,every mark/.append style={fill=teal!80!white,solid},mark=triangle\\
orange,line width=1.5pt,dashed,every mark/.append style={fill=orange!80!white,solid},mark=square*\\
Purple!60!black,line width=1.5pt,dashdotted,every mark/.append style={fill=Purple!80!white,solid},line width=1.5pt,mark=pentagon\\
WildStrawberry,line width=1.5pt,solid,every mark/.append style={solid,fill=WildStrawberry!10!white,solid},mark=otimes*\\
}

\pgfplotscreateplotcyclelist{seg_cycle}{
lime!60!black,line width=1.5pt,dashed,every mark/.append style={fill=lime!40!black,solid},mark=diamond*\\
Cyan!30!black,line width=1.5pt,dashdotted,every mark/.append style={solid,fill=Cyan!60!black},mark=pentagon*\\
WildStrawberry,solid,line width=1.5pt,every mark/.append style={solid,fill=WildStrawberry!10!white,solid},mark=otimes*\\
Mahogany,every mark/.append style={solid,fill=gray!60!white},mark=otimes*\\
blue!80!black,dashed,mark=star,every mark/.append style=solid\\
Cyan,line width=1pt,solid,every mark/.append style={fill=black,solid},mark=none\\
red,dashed,every mark/.append style={solid,fill=red!80!white,solid},mark=pentagon\\
OliveGreen!60!black,every mark/.append style={solid,fill=OliveGreen!80!white},mark=triangle\\
}

\title{Label Noise Robust Image Representation Learning based on Supervised Variational Autoencoders in Remote Sensing}
\name{Gencer Sumbul{\normalfont\textsuperscript{1}} and Beg\"{u}m Demir{\normalfont\textsuperscript{1,2}}}
\address{\textsuperscript{1}Faculty of Electrical Engineering and Computer Science, Technische Universit\"at Berlin, Germany\\
\textsuperscript{2}BIFOLD - Berlin Institute for the Foundations of Learning and Data, Germany}

\begin{document}
%
\maketitle
\begin{abstract}
Due to the publicly available thematic maps and crowd-sourced data, remote sensing (RS) image annotations can be gathered at zero cost for training deep neural networks (DNNs). However, such annotation sources may increase the risk of including noisy labels in training data, leading to inaccurate RS image representation learning (IRL). To address this issue, in this paper we propose a label noise robust IRL method that aims to prevent the interference of noisy labels on IRL, independently from the learning task being considered in RS. To this end, the proposed method combines a supervised variational autoencoder (SVAE) with any kind of DNN. This is achieved by defining variational generative process based on image features. This allows us to define the importance of each training sample for IRL based on the loss values acquired from the SVAE and the task head of the considered DNN. Then, the proposed method imposes lower importance to images with noisy labels, while giving higher importance to those with correct labels during IRL. Experimental results show the effectiveness of the proposed method when compared to well-known label noise robust IRL methods applied to RS images. The code of the proposed method is publicly available at \url{https://git.tu-berlin.de/rsim/RS-IRL-SVAE}.
\end{abstract}
\begin{keywords}
Representation learning, label noise, variational autoencoders, deep learning, remote sensing.
\end{keywords}
\section{Introduction}

The development of deep learning (DL) based remote sensing (RS) image representation learning (IRL) methods has recently gained increasing attention in the context of different learning tasks such as multi-label image classification~\cite{Sumbul:2020}, semantic segmentation~\cite{Ding:2022}, image captioning~\cite{Cheng:2022}, change detection~\cite{Zhang:2021} and content-based image retrieval~\cite{Sumbul:2022}, etc. Most of the existing IRL methods in RS require the collection of a high quantity and quality of training images annotated with pixel or scene-level labels. Gathering such data can be time-consuming and costly. As an alternative, publicly available thematic maps, automatic labeling procedures or crowdsourced data can be used as an annotation source at zero cost. 
However, this may result in including noisy labels in training data if the considered data sources include outdated information or annotation errors. Training DL-based IRL methods on such data may lead to overfitting of the considered deep neural network (DNN) on noisy labels, and thus inaccurate RS image characterization during training~\cite{Song:2022}. To address this problem, a few methods have been recently proposed in RS, aiming to achieve label noise robust IRL for semantic segmentation~\cite{ahmed_dense_2021,Dong:2022} and scene-level single/multi-label image classification~\cite{zhang_remote_2020, kang_robust_2021, li_improved_2022, Burgert:2022, aksoy_consensual_2022}. As an example, for scene-level image classification a down-weighting factor is added to softmax loss in~\cite{kang_robust_2021} to prevent the effect of images with wrong predictions (which are considered as images associated with single-label noise) on the model parameter updates of DNNs. For images associated with multi-labels, a collaborative learning framework that simultaneously operates two convolutional neural networks (CNNs) is introduced in~\cite{aksoy_consensual_2022} to detect and eliminate images with noisy multi-labels during training. In this framework, collaborative CNNs are penalized for learning similar image representations for same class predictions. Related to semantic segmentation problems, an online noise correction approach is introduced in~\cite{Dong:2022} to identify and correct pixel-level noisy labels based on information entropy during early stages of DNN training. Although these methods are potentially effective for IRL under noisy labels, each of them is designed for a specific learning problem. Their adaptation to different IRL scenarios in RS can be complex and not always feasible. 

To address the above-mentioned issues, in this paper we introduce a label noise robust IRL method that is independent from the learning problem being considered in RS. The proposed method prevents the interference of noisy labels during training, and thus accurately learns RS image representations under label noise.

\begin{figure*}[t]
  \centering
  \input{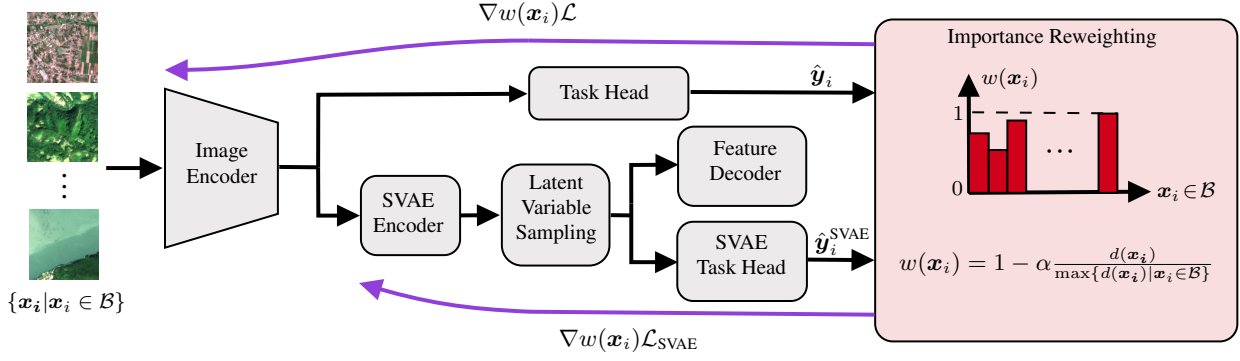}
  \caption{An illustration of the proposed label noise robust image representation learning method.}
  \label{fig:model}
\end{figure*}
\section{Proposed Label Noise Robust Image Representation Learning Method}
Let $\mathcal{T} = \{(\boldsymbol{x}_i, \boldsymbol{y}_i)\}_{i=1}^{M}$ be a training set, where $\boldsymbol{x}_i$ is the $i$th training sample and $\boldsymbol{y}_i$ is any kind of annotation asscoiated with $\boldsymbol{x}_i$. Let $\phi$ be an image encoder (e.g., CNN) that maps $\boldsymbol{x}_i$ to the corresponding representation $\boldsymbol{f}_i$ and $\psi$ be a task head that maps $\boldsymbol{f}_i$ to the corresponding label prediction $\hat{\boldsymbol{y}}_i$. A DL-based IRL problem can be formulated as finding optimum model parameters for $\phi$ and $\psi$ by minimizing the loss function $\mathcal{L}(\hat{\boldsymbol{y}}_i, \boldsymbol{y}_i)$ on $\mathcal{T}$ for a given learning task. 

The proposed method aims to achieve label noise robust IRL independently from the IRL problem. To this end, it combines a supervised variational autoencoder (SVAE)~\cite{Kingma:2014} with any kinds of $\phi$ and $\psi$ by defining variational generative process based on image features. The considered SVAE models the joint data distribution of RS image representations and the corresponding annotations by assuming that: i) $\boldsymbol{f}_i$ and $\boldsymbol{y}_i$ are generated through a latent variable $\boldsymbol{z}_i$ of the SVAE; and ii) the true posterior distribution of the latent variable is approximated with a variational approximate posterior. We define the variational approximate posterior as a multivariate Gaussian distribution, which is used to sample the latent variable as $\boldsymbol{z}_i \sim \mathcal{N}(\boldsymbol{\mu}_i,\boldsymbol{\sigma}_i^2\mathbf{I})$. Accordingly, the considered SVAE embodies a variational encoder (which is branched out from $\phi$) followed by a feature decoder and the same task head as $\psi$ with different parameters (denoted as $\psi^{\text{SVAE}}$). The feature decoder and $\psi^{\text{SVAE}}$ both form the variational decoder. The variational encoder maps $\boldsymbol{f}_i$ into the parameters $\boldsymbol{\mu}_i$ and $\boldsymbol{\sigma}_i$ of the Gaussian distribution. This allows to sample the latent variable by utilizing the reparameterization trick~\cite{Kingma:2014} as $\boldsymbol{z}_i = \boldsymbol{\mu}_i + \boldsymbol{\sigma}_i \cdot \boldsymbol{\epsilon}_i;\;\boldsymbol{\epsilon}_i\sim \mathcal{N}(\boldsymbol{0},\mathbf{I})$. Then, $\psi^{\text{SVAE}}$ produces a label prediction $\hat{\boldsymbol{y}}_i^{\text{SVAE}}$, while the feature decoder produces the reconstructed image representation $\hat{\boldsymbol{f}}_i$ based on $\boldsymbol{z}_i$. The parameters of the SVAE can be learned by maximizing the evidence lower bound (ELBO), which is defined for variational autoencoders in \cite{Kingma:2014}. To this end, for $\boldsymbol{x}_i$ we define the loss function associated with SVAE as follows:
\begin{equation}
\begin{aligned}
    \mathcal{L}_{\text{SVAE}} &= \mathcal{L}_{\text{MSE}}(\hat{\boldsymbol{f}}_i,\boldsymbol{f}_i) + \mathcal{L}(\hat{\boldsymbol{y}}_i^{\text{SVAE}},\boldsymbol{y}_i)\\&+\frac{1}{2}\sum_{j=1}^J \big(1+\text{log}(\boldsymbol{\sigma}_{i,j}^2)-\boldsymbol{\mu}_{i,j}^2-\boldsymbol{\sigma}_{i,j}^2\big),
\end{aligned}
\end{equation}
where $\boldsymbol{\mu}_{i,j}$ and $\boldsymbol{\sigma}_{i,j}$ are the $j$th element of the vectors $\boldsymbol{\mu}_{i}$ and $\boldsymbol{\sigma}_{i}$, respectively, while $J$ denotes their length. For the definition of the ELBO as a loss function, the reader is referred to~\cite{Kingma:2014}.

It is worth noting that SVAEs are less dependent on image annotations compared to non-generative DNNs~\cite{Lee:2019}. Accordingly, the loss values of training samples with noisy labels obtained through $\psi$ are expected to be relatively higher compared to those obtained through $\psi^{\text{SVAE}}$. This allows us to define the importance of each training sample for learning the model parameters based on the loss values acquired from $\psi$ and $\psi^{\text{SVAE}}$. To this end, we define our importance reweighting strategy as follows. While updating the model parameters, less importance is given to training samples associated to higher loss values through $\psi$ compared to $\psi^{\text{SVAE}}$, since these samples are considered to be associated with noisy labels. This is achieved by reweighting each training sample $\boldsymbol{x}_i$ in a given mini-batch $\mathcal{B}$ through a function $w$ as follows:
\begin{equation}
d(\boldsymbol{x_i}) = \max\{R(\mathcal{L}(\hat{\boldsymbol{y}}_i, \boldsymbol{y}_i))-R(\mathcal{L}(\hat{\boldsymbol{y}}_i^{\text{SVAE}}, \boldsymbol{y}_i)), 0\},
\end{equation}
\begin{equation}
w(\boldsymbol{x}_i) = 1 - \alpha \frac{d(\boldsymbol{x_i})}{\max\{d(\boldsymbol{x_i})|\boldsymbol{x}_i \in \mathcal{B}\}},
\end{equation}
where $\alpha$ is subject to exponential decay from 1 to 0 throughout training and $R(\boldsymbol{a})$ applies min-max rescaling on $\boldsymbol{a}$. After defining importance scores, the SVAE parameters and the remaining parameters of our method can be updated by reweighting $\mathcal{L}_{\text{SVAE}}$ and $\mathcal{L}$, respectively, with $w(\boldsymbol{x}_i)$ for each $\boldsymbol{x}_i \in \mathcal{B}$. Due to the proposed importance reweighting strategy, the IRL is achieved by mostly relying on training samples with correct labels. This leads to learning RS image representations robust to label noise independently from the considered learning problem, which can involve any kind of annotation, loss functions, $\phi$ and $\psi$. 

\section{Experimental Results}
Experiments were conducted on the BigEarthNet-S2 benchmark archive~\cite{BigEarthNet-S2}. Each Sentinel-2 image in BigEarthNet-S2 has been annotated with multi-labels from the 2018 CORINE Land Cover (CLC) database based on the 19 classes nomenclature~\cite{BigEarthNet-S2}. In addition, we have also used the CLC land cover map (pixel-level map) of each image based on the 19 classes nomenclature for the use of $\mathcal{L}$, which requires the availability of land-cover maps. For the experiments, we used the 14,832 images acquired over Serbia in summer. We divided these images into training (52\%), validation (24\%) and test (24\%) sets. To assess the robustness of our method to label noise, we injected synthetic label noise (SLN) to the training set at different rates in the range of $[10\%,60\%]$. 

In this paper, we assess our method in the context of multi-label image classification and semantic segmentation, for which cross entropy loss (CEL) function was used with scene-level and pixel-level labels, respectively. The task head is selected as an FC layer for multi-label classification, while it consists of three transposed convolutional layers (with the filters of 64, 32 and 19) for semantic segmentation. For both IRL task, the DenseNet-121 architecture~\cite{Huang:2017} was chosen as the image encoder and the latent dimension of 128 was considered for the SVAE encoder. The feature decoder of SVAE consists of an FC layer with the hidden unit size of image representation dimension (which is 1024 for DenseNet-121) or a convolutional layer with the kernel size of 1$\times$1 depending on the IRL task. We trained our method for 100 epochs by using the Adam optimizer with the initial learning rate of $10^{-3}$.

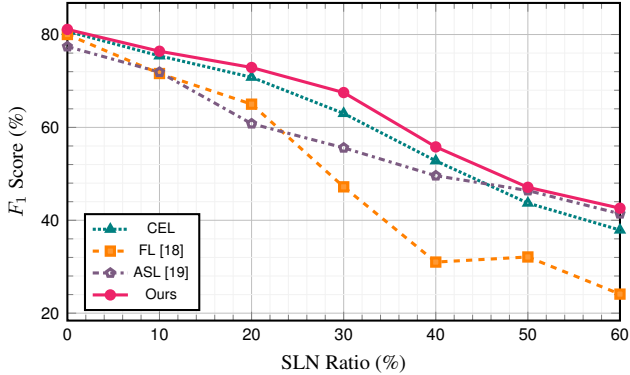
\begin{figure}[t]
  \centering
\begin{tikzpicture}[scale = 0.78]
	\begin{axis}[
    height=7cm,
    width=11cm,
    line width=1pt,
    grid=both,
    grid style={line width=.1pt, draw=gray!10},
    major grid style={line width=.2pt,draw=gray!50},
    legend pos=south west,
    minor x tick num=4,
    minor y tick num=4,
    xlabel= {\normalsize SLN Ratio (\%)},
    ylabel= {\normalsize $F_1$ Score~(\%)},
    xmin=0,xmax=60,
    cycle list name=cls_cycle]
    \addplot+[name path=capacity] table [x=noise_prcnt, y=Cls, col sep=comma] {BEN_results.csv};\addlegendentry{CEL};
    \addplot+[name path=capacity] table [x=noise_prcnt, y=Focal, col sep=comma] {BEN_results.csv};\addlegendentry{FL~\cite{Lin:2020}}
    \addplot+[name path=capacity] table [x=noise_prcnt, y=AsymmetricLoss, col sep=comma] {BEN_results.csv};\addlegendentry{ASL~\cite{Ridnik:2021}};    
    \addplot+[name path=capacity] table [x=noise_prcnt, y=OurCls, col sep=comma] {BEN_results.csv};\addlegendentry{Ours};
    \end{axis}
    \end{tikzpicture}
  \caption{$F_1$-score versus SLN ratio for multi-label RS image classification.}
  \label{fig:cls_res}
\end{figure}

\begin{figure}[t]
  \centering
\begin{tikzpicture}[scale = 0.78]
	\begin{axis}[
    height=7cm,
    width=11cm,
    grid=both,
    grid style={line width=.1pt, draw=gray!10},
    major grid style={line width=.2pt,draw=gray!50},
    legend pos=south west,    
    minor x tick num=4,
    minor y tick num=4,
    xlabel= {\normalsize SLN Ratio (\%)},
    ylabel= {\normalsize OA~(\%)},
    xmin=0,xmax=60,
    cycle list name=seg_cycle]
    \addplot+[name path=capacity] table [x=noise_prcnt, y=Seg, col sep=comma] {BEN_results.csv};\addlegendentry{CEL};
    \addplot+[name path=capacity] table [x=noise_prcnt, y=LearningWithNoiseCorrection, col sep=comma] {BEN_results.csv};\addlegendentry{LWNC~\cite{Dong:2022}};
    \addplot+[name path=capacity] table [x=noise_prcnt, y=OurSeg, col sep=comma] {BEN_results.csv};\addlegendentry{Ours};
    \end{axis}
    \end{tikzpicture}
  \caption{Overall accuracy (OA) versus SLN ratio for semantic segmentation of RS images.}
  \label{fig:seg_res}
\end{figure}
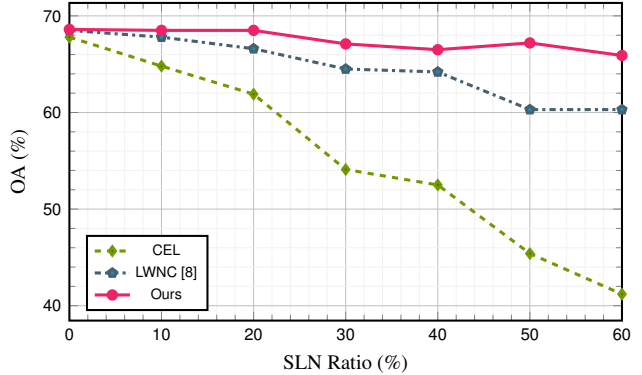
We compared our method with IRL through focal loss (denoted as FL)~\cite{Lin:2020}, asymmetric loss (denoted as ASL)~\cite{Ridnik:2021} and the binary CEL for multi-label image classification in terms of $F_1$ score. It was also compared with the high-resolution land cover mapping through learning with noise correction method (denoted as LWNC)~\cite{Dong:2022} and IRL through the pixel-wise CEL for semantic segmentation in terms of overall accuracy (OA). Figures~\ref{fig:cls_res} and \ref{fig:seg_res} show the corresponding results. By assessing the figures, one can see that our method leads to highest scores compared to other methods for all SLN ratios in the context of both multi-label image classification and semantic segmentation. As an example, the proposed method achieves 24\% higher $F_1$ score compared to FL in the context of multi-label image classification when SLN ratio is 40\% (see Fig.~2). In the context of semantic segmentation, our method provides 6\% higher overall accuracy compared to LWNC when 60\% of labels in the training data are noisy (see Fig.~3). These results show that our method accurately learns deep representations of RS images when training data includes pixel-level or scene-level noisy labels. This is due to the effective integration of SVAEs into non-generative DNNs that allows to define the importance of each training sample and to reweight them accordingly during IRL. One can also observe from the figures that the performance of the proposed method is less affected by the increase in SLN ratio compared to other methods. This is more evident for the IRL task of semantic segmentation, which is a relatively more complex task than multi-label classification and can be highly affected by the interference of noisy pixel-level labels. As an example, when the SLN ratio is increased to 60\% from 0\%, IRL through the pixel-wise CEL results in almost 25\% more decrease in overall accuracy compared to our method. This shows that our method achieves label noise robust IRL under high noise ratios more accurately than the other methods. All these results under two different learning tasks with the corresponding DNN architectures and two different image annotation types show that the proposed method can be effectively employed for different IRL problems.



\section{Conclusion}
In this paper, we have presented a novel label noise robust IRL method that can be applied to any IRL problem in RS. The effectiveness of our method relies on its capability to: i) combine SVAEs with any kind of DNNs that allows to define importance of training samples during IRL; and ii) accurately learn RS image representations under label noise by imposing lower importance to training samples with noisy labels. Experimental results show the success of our method compared to well-known label noise robust IRL methods. Although the experiments were conducted in the context of multi-label classification and semantic segmentation of RS images under scene-level and pixel-level noisy labels, our method can be applied to any IRL task, loss function, DNN architecture or annotation type. This can be a very crucial advantage in large-scale RS applications that may require to learn RS image representations under different IRL scenarios. 

We would like to note that the training samples, which are associated with a very high label noise, may continue to interfere IRL training even though a low importance is given to these samples in our method. Thus, these samples may require to be eliminated from the considered training set. As a future development of this work, we plan to integrate our method with unsupervised IRL of such samples. This can allow to further prevent the interference of label noise without reducing the training set size.

{\section{Acknowledgments}}
This work is supported by the European Research Council (ERC) through the ERC-2017-STG BigEarth Project under Grant 759764, and by the German Research Foundation through the IDEAL-VGI project under Grant 424966858, and by the European Space Agency (ESA) through the Demonstrator Precursor Digital Assistant Interface For Digital Twin Earth (DA4DTE) Project.
\vfill
\bibliographystyle{IEEEtran}
{\small
\bibliography{refs}}
\vfill
\end{document}